\pdfoutput=1
\documentclass{article}
\PassOptionsToPackage{numbers, compress}{natbib}
\usepackage[nonatbib,preprint]{neurips_2019}

\usepackage[numbers]{natbib}

\usepackage[utf8]{inputenc} 
\usepackage[T1]{fontenc}    
\usepackage{hyperref}       
\usepackage{url}            
\usepackage{booktabs}       
\usepackage{amsfonts}       
\usepackage{nicefrac}       
\usepackage{microtype}      
\usepackage[utf8]{inputenc}
\usepackage[english]{babel}
\usepackage{multirow}
\usepackage{float}

\usepackage{mathrsfs}
\usepackage{amsmath,amssymb,bm}
\usepackage{graphicx}
\usepackage{algorithm}
\usepackage{algorithmic}
\usepackage{arydshln}
\usepackage{multicol}
\usepackage{verbatim}
\usepackage{tikz}
\usepackage{longtable}

\title{Automatic coding of students’ writing via Contrastive Representation Learning in the Wasserstein space}

\author{%
  Ruijie Jiang, Julia Gouvea, David Hammer, Shuchin Aeron \thanks{Research supported by NSF 1931978. S. Aeron is also supported by NSF CCF:1553075, NSF TRIPODS grant HDR:1934553, and AFOSR FA9550-18-1-0465. emails: ruijie.jiang@tufts.edu, julia.gouvea@tufts.edu, david.hammer@tufts.edu, shuchin@ece.tufts.edu} \\
  Tufts Univerity\\
  Medford, MA -2155
}

\begin{document}

\maketitle

\begin{abstract}
Qualitative analysis of verbal data is of central importance in the learning sciences. It is labor-intensive and time-consuming, however, which limits the amount of data researchers can include in studies. This work is a step towards building a statistical machine learning (ML) method for achieving an automated support for qualitative analyses of students’ writing, here specifically in score laboratory reports in introductory biology for sophistication of argumentation and reasoning. We start with a set of lab reports from an undergraduate biology course, scored by a four-level scheme that considers the complexity of argument structure, the scope of evidence, and the care and nuance of conclusions. Using this set of labeled data, we show that a popular natural language modeling processing pipeline, namely vector representation of words, a.k.a word embeddings, followed by Long Short Term Memory (LSTM) model for capturing language generation as a state-space model, is able to quantitatively capture the scoring, with a high Quadratic Weighted Kappa (QWK) prediction score, when trained in via a novel contrastive learning set-up. We show that the ML algorithm approached the inter-rater reliability of human analysis. Ultimately, we conclude, that machine learning (ML) for natural language processing (NLP) holds promise for assisting learning sciences researchers in conducting qualitative studies at much larger scales than is currently possible. 

\end{abstract}

\section{Introduction}
In recent years, science, engineering and mathematics education has emphasized supporting students’ disciplinary "practices" of inquiry~\cite{ford2015educational,national2012framework}. These practices—such as of formulating questions, designing investigations, or arguing from evidence—are more difficult to identify and assess than the traditional objectives of particular correct content knowledge. In order to study students’ practices, researchers rely mainly on qualitative analyses of naturalistic data~\cite{manz2012understanding,metz2019primary,lehrer2015development}. These studies have advanced the field’s understanding of practices.\\

These studies have been limited in scope, however, because they are extremely labor-intensive: Analysis of naturalistic data requires significant and extensive effort by trained researchers, from transcribing to coding to the construction of meaning. It has been time and cost-prohibitive to conduct qualitative studies with large samples of data. Our purpose in this project is to develop computational tools that can support qualitative research at large scales on students' inquiry practices in science. In this paper we report on initial progress towards applying natural language processing (NLP) techniques to research on students' written arguments in college biology laboratory reports. \\

In this work, we report on our success in designing NLP that approached the reliability of trained, human coders. Specifically, we show that contrastive learning in the Wasserstein space is able to achieve a high level of agreement (as measured by QWK) on average. \\

The rest of the paper is organized as follows. In section \ref{sec:NLPinLS} first overview current state-of-art in automating assessment of writing in science using machine learning for natural language processing. Following that we outline the writing assessment setting particular to our case in section \ref{sec:bio}.  In section \ref{sec:approach} we briefly survey the relevant literature in machine learning for NLP and introduce our novel approach for automatic scoring. In section \ref{sec:exp} we evaluate the performance of the proposed approach and discuss the results in detail.\\

\section{Examining science writing using NLP machine learning approaches}
\label{sec:NLPinLS}
The most common use of ML/NLP in science education has been to assess the conceptual correctness of students' written explanations or arguments~\cite{zhai2020applying}. For example, Nehm and colleagues~\cite{nehm2012transforming} have used ML techniques to assess students' explanations of evolutionary change. They and other researchers have found that ML can match the reliability of human coders in scoring or categorizing written explanations~\cite{nehm2012transforming,ha2011applying,liu2016validation}. \\

These results are encouraging for the general goal of conducting qualitative research at scale and they provide a useful complement to existing large-scale methods of assessing conceptual understanding. The same algorithms, however, cannot apply to research on students’ inquiry practices. Assessing conceptual correctness can rely on using ML to recognize a limited set of target terms or phrases. For example, explanations of evolutionary change should include ideas about variation, heritability and selection~\cite{ha2011applying}. Typically, responses will receive a higher score if they contain more of these target ideas and fewer incorrect or irrelevant ideas. Computational algorithms can learn to code for the presence or absence of these targets fairly easily. For researchers interested in expressions of scientific practice, that may contain novel ideas, to be effective the ML approach cannot rely on specific target terms.\\ 

To study these aspects of scientific practice, researchers look for evidence of how students support, relate and evaluate scientific claims~\cite{kelly2002epistemic}. Interpreting the disciplinary value of students' scientific writing requires more attention to context and the ability to evaluate the value of novel or unexpected ideas. This creates a new set of challenges for NLP/ML and raises questions about the ability of ML to deal with the more creative aspects of scientific practice.\\ 

Lee and colleagues~\cite{lee2019automated} have made progress in this direction by using NLP/ML to score the quality of students’ uncertainty-infused scientific arguments. Lee~\cite{lee2019automated} scored students ’ explanations environmental phenomena according to the quality of evidence and reasoning. They also scored students’ reflections on the certainty of their explanations. Using an NLP engine called c-rater-ML, they found relatively high human-machine agreement (QWK > 0.70) on both constructs. Rather than identify pre-specified target words, c-rater-ML extracts features of text including characters, word sequences and word associations to model responses, and then uses support-vector methods to make score predictions. This work suggests that automated techniques can be successful in detecting qualitative markers of sophisticated science writing in text.\\

Building on this work, our aim was to explore the possibility that NLP techniques could identify markers of quality in university students' laboratory reports. In addition, our intention in using ML was to explore its role in aiding researchers. ML has commonly been positioned as a substitute for an instructor or grader, making it possible to provide feedback or grades to large numbers of students. Our aim has been to explore the utility of ML as a research tool that can allow researchers to detect patterns and consider meaning in qualitative data.

\section{Complexity of argumentation in biology lab reports}
\label{sec:bio}

The present work concerns analyses by (Gouvea et al. in prep) to track shifts in the quality of students’ writing due to curricular innovations. (Gouvea et al. in prep) adapted the domain-general Structure of Observed Learning Outcomes (SOLO) taxonomy~\cite{biggs2014evaluating} to study the quality of argumentation in college biology lab reports. The scheme defines quality in terms of three features of written arguments:
\begin{enumerate}
    \item \textbf{Argument structure} refers to the number and inter-relatedness of claims.
    \item \textbf{Scope of knowledge/evidence} refers to the sources of evidence used the degree to which that evidence is reconciled with other forms of evidence.
    \item \textbf{Consistency \& closure} refers to the extent to which conclusions are theoretically consistent and appropriately closed or left open in the face of uncertainty.
\end{enumerate}

These features inform the choice of a score on a four-level scale. Table 1 describes and provides examples of student writing for each of four levels within each category. For example, in the category of argument structure, Level 1 was assigned to tautological arguments that repeated back given facts as claims. Level 2 was assigned to claims that made one-sided inferences. Level 3 was assigned to arguments that considered at least two possible claims, but ultimately sided with one claim. Finally, Level 4 arguments integrated claims to make compound or conditional arguments. \\
In practice, it was not always possible for coders to clearly assign a level to each of the three categories. Instead, coders evaluated each response according to each category as much as possible and then assigned an overall score of 1-4. 

\begin{table}[]
\begin{tabular}{|l|l|l|l|l|}
\hline
 & Level 1 & Level 2 & Level 3  & Level 4 \\ \hline
\begin{tabular}[c]{@{}l@{}}Argument\\  Structure\end{tabular}            & \begin{tabular}[c]{@{}l@{}}Tautological/\\ Simple\end{tabular}                                                  & \begin{tabular}[c]{@{}l@{}}Additive, \\ One-sided\end{tabular}                                                 & \begin{tabular}[c]{@{}l@{}}Relational,\\ Two-sided\end{tabular}                                                                      & \begin{tabular}[c]{@{}l@{}}Compound/\\ Conditional\end{tabular}                                                                                                               \\ \hline
\begin{tabular}[c]{@{}l@{}}Scope of\\ Knowledge/\\ Evidence\end{tabular} & \begin{tabular}[c]{@{}l@{}}Relies on\\ given information\\ and/or vague \\ reference \\ to results\end{tabular} & \begin{tabular}[c]{@{}l@{}}Uses subset of\\ available data as \\ evidence to support \\ inference\end{tabular} & \begin{tabular}[c]{@{}l@{}}Uses more of\\ the data (including \\ contradictory \\ patterns) \\ to support \\ inferences\end{tabular} & \begin{tabular}[c]{@{}l@{}}Uses data (including \\ contradictory patterns) \\ as well as\\ outside  information  \\ or  hypotheticals/  \\ thought  experiments\end{tabular} \\ \hline
\begin{tabular}[c]{@{}l@{}}Consistency\\ \& Closure\end{tabular}         & \begin{tabular}[c]{@{}l@{}}Closed\\ without rationale, \\ ignores \\ inconsistencies\end{tabular}               & \begin{tabular}[c]{@{}l@{}}Closed\\ with some rationale,   \\ ignores \\ inconsistencies\end{tabular}          & \begin{tabular}[c]{@{}l@{}}Closed, addresses\\ and reconciles \\ inconsistencies\end{tabular}                                        & \begin{tabular}[c]{@{}l@{}}Open-ended or\\ appropriately \\ qualified\end{tabular}                                                                                            \\ \hline
\end{tabular}
\caption{Description of dimensions that comprise each level.}
\end{table}

\subsection{Data description}
This data set is composed of the Discussion sections of 146 lab reports from a Biology course. The reports were scored by human education researchers using a rubric adapted from Biggs and Collis~\cite{biggs2014evaluating}. The properties of the lab reports are shown in Table 2.

\begin{table}[h!]
\begin{center}
\begin{tabular}{|c|c|c|c|c|} 
\hline
Essays & Avg length & Max length & Min length & Scores\\
\hline
146 & 512 & 1449 & 95 & 1-4\\ 
\hline
\end{tabular}
\end{center}
\caption{ Statistics of the dataset}
\end{table}

In this study we explore the ability of NLP techniques to differentiate among samples of students’ scientific writing by Gouvea et al s scheme. The target is to match human-human inter-rater reliability.

\section{Proposed Approach}
\label{sec:approach}

Our approach departs from existing ML work in examining science writings as surveyed previously and heavily based on using numerical representations or vector embeddings of words and documents, popularly called as word2vec~\cite{mikolov2013distributed} or doc2vec~\cite{le2014distributed}, which capture lexical and semantic properties of words. These methods employ a self-supervised learning mechanism to create universally useful representations that have shown to be useful for tasks such as sentiment analysis, topic segmentation. With these word embeddings/representations, both Recursive Neural Networks (RNN) and Convolution Neural Networks (CNN) can then be used for modeling the input document. It has been commonly understood that CNN can capture the local information~\cite{zhao2020cloud} and RNN can capture the sequential information, capable of modeling long term dependencies using architecture such as Long-Short-Term-Memory (LSTM)~\cite{hochreiter1997long}, bi-LSTM~\cite{schuster1997bidirectional}. These models are further refined to handle longer-range dependencies, e.g. contextual LSTM (CLSTM)~\cite{ghosh2016contextual} and attention-based LSTM~\cite{tang2016aspect}. A relevant work that uses vector embeddings for Automatic Essay Scoring (AES) is Kaveh~\cite{taghipour2016neural}, where the authors employed RNN for automated scoring. However, as we show below the approach in Kaveh~\cite{taghipour2016neural} does not capture all the temporal information from the RNN and therefore suffers from low accuracy, especially in cases with low amounts of training data.\\

In our context, the prime use of these embeddings as well as the subsequent use of CNN and RNN as encoders, is to measure similarity between documents, and using this similarity employ the contrastive representation learning with triplets loss \cite{oord2018representation, hermans2017defense} for learning discriminative representations, across different classes. As a similarity measure we propose to use the Word Mover’s Distance \cite{kusner2015word}), which is the Wasserstein distance \cite{cuturi2013sinkhorn,COT_book} between document representations, and develop the novel approach of contrastive document representation learning in the Wasserstein space. Embedding natural language data into Wasserstein spaces have recently been shown to be useful for NLP tasks~\cite{frogner2019learning}, and we are motivated by these developments. 
We next describe the proposed workflow in detail.

\subsection{ Proposed workflow for automatic scoring}

\begin{figure}[htb]
\begin{center}
\includegraphics[width=14cm, height=2cm]{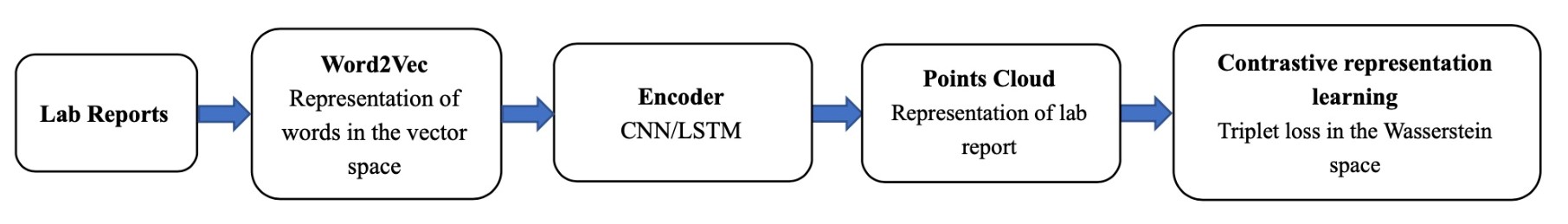}
\end{center}
\caption{Overview of the main workflow – Contrastive Representation Learning in Wasserstein Space (CoReL-W).}
\label{fig:NB}
\end{figure}
Figure 1. shows the main workflow for Contrastive Representation Learning in Wasserstein Space (CoReL-W) and consists of the following modules.\\

\textbf{Word Representation}: The lab report is modeled as a sequence of words $\bm w_t$, and each word, in turn, is represented by a vector $\bm x_t \in \mathbb{R}^{d_w}$. So, a report with N words is represented by a matrix $\bm X = \bm{(x_1, x_2,...x_N)} \in \mathbb{R}^{d_w \times N}$.\\

\textbf{Encoder}: This matrix $\bm X$ is then fed into an encoder module, that computes a representation of the entire document. We use two different encoders in our model, viz. Convolution Neural Network (CNN) and Recurrent Neural Network (RNN), and we describe the details below. 

\begin{itemize}
\item \textbf{Convolution Neural Network (CNN)}: In order to extract local features from the sequence~\cite{zhao2020cloud}, a shared linear transformation matrix $\bm S$ is applied to all windows. We define $2l+1$ as the window size. The words in the window can be represented as a vector $\bm x_{tl} = (\bm x_{t-l}^\top, \bm x_{t-l+1}^\top,...,\bm x_{i+l}^\top) \in \mathbb{R}^{(2l+1).d_w}$. For matrix $\bm{S} \in \mathbb{R}^{(2l+1).d_w \times d_c}$ and bias $\bm g \in \mathbb{R}^{d_c}$, the output vector $\widetilde{\bm x}_t \in \mathbb{R}^{d_c}$:
\begin{equation}
\widetilde{\bm{x}}_t = \bm{x}_{tl}\bm S + \bm g
\end{equation}
To guarantee the number of words within the window remains the same, we did zero padding for each document. The output of this encoder is the set $\widetilde{\mathcal{X}} =\{ \widetilde{\bm x}_1, \widetilde{\bm x}_2,..., \widetilde{\bm x}_N\}$, which can be viewed as the representation of the lab report for downstream task.\\
\begin{figure}[H]
\begin{center}
\includegraphics[width=10cm, height=5cm]{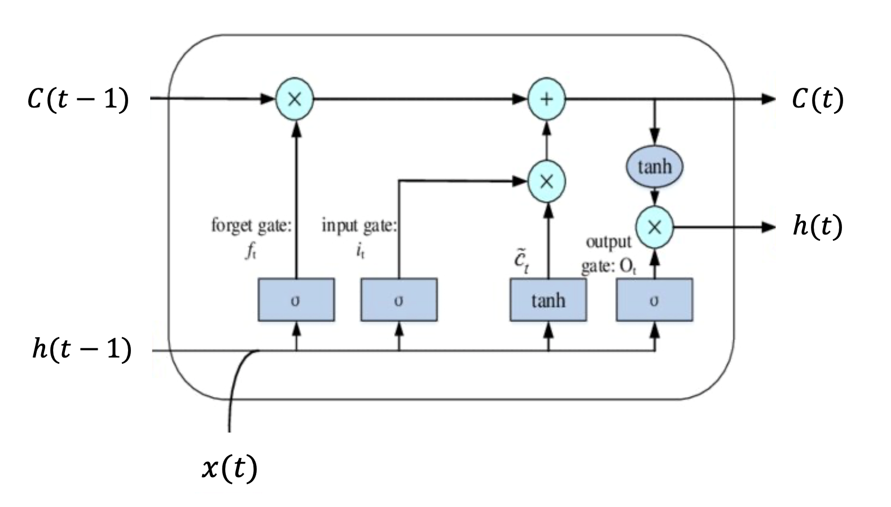}
\end{center}
\caption{The LSTM architecture~\cite{yuan2019nonlinear}}
\label{fig:NB}
\end{figure}

\item \textbf{Recurrent Neural Network (RNN)}: The RNN architecture that we will use here is the standard LSTM~\cite{hochreiter1997long} architecture. The structure of a LSTM model is shown in Figure 2., and the Equations 2 describe how LSTM works:
\begin{equation} 
\label{eq:LSTM}
\begin{split}
\bm i_t = \sigma({\bm x}_{t} \bm U_i+\bm h_{t-1} \bm W_i+\bm b_i),    \\
\bm f_t  = \sigma({\bm x}_{t} \bm U_f+\bm h_{t-1} \bm W_f+\bm b_f),\\
\bm o_t = \sigma({\bm x}_{t} \bm U_o+\bm h_{t-1} \bm W_o+\bm b_o),\\
\bm{\widetilde{\bm C}_t} = tanh({\bm x}_{t} \bm U_g+\bm h_{t-1} \bm W_g+\bm b_c),\\
\bm C_t  = \sigma(\bm f_{t} \circ \bm C_{t-1}+\bm i_{t} \circ  \widetilde{\bm C}_t),\\
\bm h_t = tanh(\bm C_t) \circ \bm o_t.
\end{split}
\end{equation}
$\sigma$ represent the sigmoid activation function $\sigma(r) =  \dfrac{1}{1 + e^{-r} }$, and $tanh$ represent the tanh activation function $tanh(r) = \dfrac{e^{2r} - 1}{ e^{2r} + 1}$:\\

The set of Equations \eqref{eq:LSTM} is depicted as a unit in Figure \ref{fig:NB}. For each unit, the input contains three parts, ${\bm x}_t,\bm h_{t-1},\bm C_{t-1}$. ${\bm x}_t$ corresponds to the $t_{th}$ word vector, $\bm h_{t-1}$ and $\bm C_{t-1}$ outputs from the previous unit. $\bm{U},\bm{W}$ are matrices that are parameters to be learned, $\bm{b}$ are learned bias vectors. The symbol $\circ$ denotes the Hadamard product. The vector $\bm h_t\in\mathbb{R}^{d_h}$ corresponds to the hidden state at time step $t$. 

The overall output of the LSTM is the set of hidden states $\mathcal{H} = \{\bm h_1, \bm h_2,..., \bm h_N \}$, which can be viewed as the representation of the lab report for downstream task. In our experiments, we also use biLSTM ~\cite{schuster1997bidirectional}, which consists of two LSTMs, one that processes the input in forward direction $\bm{(x_1, x_2,...x_N)}$, one that processes the input in forward direction  $\bm{(x_N, x_{N-1},...x_1)}$ separately.
\end{itemize}

\textbf{Learning representation in the Wasserstein space}: Given the representation of a lab report as a set of support points $\mathcal{Z} = \{\bm z_1, \bm z_2,..., \bm z_N \}$ generated from the encoder. Our objective is to tune the parameters of the encoder to learn useful representation. Note that $\mathcal{Z} = \tilde{\mathcal{X}}$ when using the CNN as the encoder and $\mathcal{Z} = {\mathcal{H}}$
when using the RNN as the encoder. Please refer to the notation in the discussion above. 

To learn good encoder representations, we propose a novel approach, namely that of contrastive learning in the Wasserstein space that we outline below. The \textbf{\emph{key idea}} is to view the output of the encoder as empirical distributions, supported on the finite sets of points $\mathcal{Z}$. For two different reports with representations $\mathcal{Z}^1 =  \{\bm z_1^1, \bm z_2^1,..., \bm z_{N_1}^1 \}$ and $\mathcal{Z}^2 = \{\bm z_1^2, \bm z_2^2,..., \bm z_{N_2}^2 \}$, we denote the corresponding empirical distribution $\mu$ and $\nu$ via,
\begin{equation}
\mu = \sum_{i=1}^{N_1} \boldsymbol{u_i} \delta_{{\bm z}_{i}^1},  \nu = \sum_{i=1}^{N_2} \boldsymbol{v_i} \delta_{{\bm z}_i^2}.
\end{equation}
Here, $\bm u$ and $\bm v$ are vectors of non-negative weights summing to 1 separately and $\delta$ is the dirac delta function. One can compare two distributions using the Wasserstein distance ~\cite{COT_book} defined as follows:
\begin{equation}
\begin{array}{rrclcl}

\mathcal W(\mu,\nu) = \inf_{\Gamma \in \mathbb{R}^{N_1 \times N_2}, \Gamma \geq 0} \, \sum_{i,j} d({\bm z}_i^1,{\bm z}_j^2) \Gamma (i,j)\\
\textrm{s.t.}  \sum\limits_{i} \Gamma (i,j) = \bm{v}_j, \sum\limits_{j} \Gamma (i,j) = \bm{u}_i, \\
\end{array}
\end{equation}
where $ d({\bm z}_i^1,{\bm z}_j^2) = \frac{1}{2}\| {\bm z}_i^1- {\bm z}_j^2\|^2$. To allow for end-to-end training, we use entropic regularization and use the Sinkhorn algorithm to compute the Wasserstein distance~\cite{cuturi2013sinkhorn}. 

Given this set-up, our approach is to learn useful representations via contrastive learning using the triplet loss ~\cite{oord2018representation, hermans2017defense}. In this approach, one defines triplets $(a,a^+,a^-)$, where $a$ is the anchor sample (report), $a^+$ is the positive sample, $a^-$ is the negative sample, and use the Wasserstein distances $\mathcal{W}(\mu_i^a, \mu_i^{a^+})$ and $\mathcal W(\mu_i^a, \mu_i^{a^-})$ as similarity measure, where $\mu^a$, $\mu^{a^+}$ and $\mu^{a^-}$ are representations of the anchor, positive sample and negative sample respectively as empirical distributions. Given $P$ such triplets, we minimize the following contrastive loss function~\cite{ hermans2017defense}:
\begin{equation}
\mathcal L(\mu^a, \mu^{a^+}, \mu^{a^-}) = \frac{1}{P} \sum_{i=1}^{P} \max(\mathcal W(\mu_i^a, \mu_i^{a^+}) - \mathcal W(\mu_i^a, \mu_i^{a^-}) + m, 0), 
\end{equation}
where the constant $m$ is a margin hyperparameter that controls the difference of the distance between positive pair and negative pair.

Noting that the distributions $\mu^a$, $\mu^{a^+}$ and $\mu^{a^-}$ are function of the parameters $\theta$ of the encoder, these parameters are learned on a training set via solving the following optimization problem.  
\begin{equation}
\theta_0  = \arg\min_{\theta}\mathcal L(\mu_\theta^a, \mu_\theta^{a^+}, \mu_\theta^{a^-})
\end{equation}

\textbf{KNN Prediction}: Given the learned parameters from a training set, say  $\theta_0$, of the encoder, each report from training set can be represented by the corresponding representation $\mu_{\theta_0}^i, i = 1,2,..., N_{train}$, where $N_{train}$ is the number of reports in the training set. To predict the score $s_{test}$ of a report from the test set, the Wasserstein distance between its representation $\mu_{\theta_0}^{test}$ and the representation of all reports from training set $\{\mu_{\theta_0}^{i}\}_{i=0}^{N_{train}}$ is calculated.  Then $K$ reports from the training set with the lowest Wasserestein distance to the test $\mu_{\theta_0}^{test}$ are selected. Let these scores be $(s_1, s_2,..., s_K)$. Then the declared score is simply the nearest integer rounding of the average score $s_{test} = round(\frac{1}{K}\sum_{i=1}^{K}s_i)$.

\section{Evaluation of the proposed approach on Biology lab report scoring}
\label{sec:exp}

\subsection{Evaluation Metric}
We use Quadratic Weighted Kappa~\cite{cohen1968weighted} to evaluate our system that is calculated as follows. First, a weight matrix $\bm{A}\in \mathbb{R}^{M \times M}$ is calculated in as follows:
\begin{equation}
\bm A_{p,q} = \frac{(p-q)^2}{(M-1)^2}
\end{equation}
where $p$ and $q$ are the true score and the predicted score respectively, and $M$ is the number of possible ratings. A matrix  $\bm O \in \mathbb{R}^{M \times M}$ is calculated such that $\bm O_{p,q}$ denotes the number of essays whose score is $p$ and prediction $q$. An expected count matrix $\bm E\in \mathbb{R}^{M \times M}$ is calculated where $E_{p,q}$ equals to the product of the number of essays whose true score is $p$ and the predicted score is $q$. The matrix $\bm E$ is then normalized such that the sum of elements in $\bm E$ and the sum of elements in $\bm O$ are the same. Finally, the QWK score is calculated as:
\begin{equation}
\mbox{QWK} = 1 - \frac{\sum_{p,q}\bm A_{p,q}\bm O_{p,q}}{\sum_{p,q}\bm A_{p,q}\bm E_{p,q}}
\end{equation}

\textbf{Selection of Triplets for contrastive learning}: For each report, we randomly choose one report with the same score as a positive sample and choose another report with different score as a negative sample. This negative sample is selected based on the following probability:
\begin{equation}
Prob(D_j) = \frac{ \left|s-s_j \right|}{\sum\limits_{k=1}^{M}{\left|s-s_k \right|}}.
\end{equation}
Here, $D_j$ is the report to be chosen, $s$ and $s_j$ are the score of the original report and the report to be chosen. $M$ is the number of possible ratings. In other words, the probability probability is proportional to the difference between the scores of the anchor and the sample to be chosen, which means the sample whose score is more different will have a higher probability to be chosen as the negative sample. In our experiment, we generate 8 triplets for each lab report in the training set. This makes the total number of triplets to be 816.

\subsection{Setup}
We used 10-fold cross-validation to evaluate our system. In each fold, 70\% of the data is used as the training set and the other 30\% as the test set. We use Quadratic Weighted Kappa to evaluate our system. A max number of epochs was set to train the model. To clean the text, we removed the punctuation and stop words, lowercase, stemming, and lemmatize the text by using NLTK~\cite{loper2002nltk}. \\

Our system has some hyper-parameters that need to be set. The optimizer we used was Adam~\cite{kingma2014adam}, the learning rate was set to 0.01. The mini-batch size was 408. The number of epochs was 5. We also employed batch normalization~\cite{ioffe2015batch} during the training process. The length of the word embedding $d_w$ was 300, the length of the hidden state for RNN $d_h$ was 300. We did zero paddings for the CNN, and the window size we set was 3, the output length $d_c$ was 300. We employed the pre-trained GLoVe~\cite{pennington2014glove} word vectors as the representation of words. Before being fed into the neural network, each word embedding was multiplied with a certain weight from referred to as the smooth inverse frequency (SIF)~\cite{arora2016simple}, this weight has a negative relevance to the frequency of the words. The parameter was initialized randomly. For prediction using the KNN, we set the number of nearest neighbors to K = 7.

\subsection{Result and Discussion}
The various methods we used for scoring are shown in Table 3. As a baseline, we also rebuild a widely used automated scoring system called “c-rater-ML”~\cite{leacock2003c}. "c-rater-ML '' is a key words and key sentences detector-based system, which has been proven to successfully score short answers automatically. As shown in Table 4, we can see "c-rater-ML'' fails in our task compared to the proposed methods. For reference, note that the QWK for reliability between human coders was estimated at 0.839. According to Cohen~\cite{cohen1968weighted}, values of QWK between 0.41 - 0.6 are considered "moderate" and values between 0.61 - 0.8 are considered "substantial." There the proposed methods achieve a good level of performance. 

\begin{table}[h!]
\begin{center}
\begin{tabular}{ |c|p{9cm}| } 
\hline
Method & Description  \\
\hline
c-rater-ML & Five features:Word n-grams, Character n-grams, Response length, Syntactic dependencies 
and Semantic role labels. Then use SVR to train the model. \\ 
LSTM & Use LSTM as an encoder, make prediction using the average of the hidden state vectors~\cite{taghipour2016neural}.  \\ 
biLSTM & Use biLSTM in Kaveh’s method~\cite{taghipour2016neural}. \\ 
CoReL-W+biLSTM &  CoReL-W with biLSTM as the encoder + KNN for prediction. \\ 
CoReL-W+CNN &  CoReL-W with CNN as the encoder + KNN for prediction. \\ 
\hline
\end{tabular}
\caption{ Methods we used in the experiment}
\end{center}
\centering 
\end{table}

\begin{table}[h!]
\begin{center}
\begin{tabular}{cc}
\hline
System & QWK\\
\hline
c-rater-ML & 0\\
LSTM &  0.46\\
biLSTM & 0.61\\
CoReL-W+biLSTM & 0.628\\
CoReL-W +CNN & 0.645\\
\hline
\end{tabular}
\caption{The performance of the various models} 
\end{center}
\end{table}

\subsection{Re-coding Process}
During the cross-validation process we evaluated the reliability of ML predictions against human scores.  ML predictions consistently matched human coders for 5 lab reports; ML predictions consistently predicted a score that differed from human coders for 9 reports (Table 5). Three of the coders who had coded the original dataset participated in a re-coding process. This re-coding took place over a year after the original coding and coders were blind to both the original code and the ML prediction. Each report was re-coded by at least two coders independently and was then discussed by all three coders until a consensus score was agreed upon. Cohen's (linear weighted) kappa amongst coders for this subset (N=14) was 0.566; while QWK was 0.701. Note that both values are lower than the kappa values for the full dataset, indicating that this subset of 14 reports may have been more challenging to code. \\

Table 5 compares the original consensus codes by human coders with the ML predictions and the consensus score from the re-coding process. Of the 9 reports for which the ML prediction consistently differed from the original human-coded consensus score, 4 were revised in re-coding in the direction of the ML prediction. The remaining 5 examples were not changed from the original. Of the 5 reports for which the ML prediction consistently matched the original human coded score, three remained matching in re-coding, while two reports were revised away from both the original code and the ML prediction.\\
It is noteworthy that the two instances in which the re-coding process resulted in revision away from the ML prediction involved reports coded as Level 1. The ML process was least successful in predicting Level 1 reports, likely because the data set was imbalanced (only 12 out of 146 were coded as Level 1 by human coders).\\

That the ML prediction matched revised scores from the re-coding process suggests the possibility that ML may be useful for identifying examples that human coders may want to revisit, either because they were coded inconsistently or because they represent difficult to code "borderline" examples.

\begin{table}
\begin{center}
\begin{tabular}{cccccc}
\hline
A & Report ID	&ID &Original score	&ML prediction	&Re-code score\\
\hline
\multirow{9}{6em}{ML prediction consistently differs from original score} 
&3BS15	&121&	1	&3	&3$^*$\\
&C10	&13	&2	&3	&2\\
&3BS16	&107&	2	&3&	2\\
&E5 &26	&2&	3&	3$^*$\\
&B7 &61	&3&	4&	4$^*$\\
&H20	&3	&4  &3   &3$^*$\\
&3AS17	&91&	4&	3&	4\\
&3BS8	&122&	4&	3&	4\\
&3BS17	&115	&4&	3&	4\\
\hline
\multirow{5}{6em}{ML prediction consistently matches original score}
&K15&	19&	2&	2&	1\\
&3As7&	89&	2&	2&	1\\
&3AS6&	90&	2&	2&	2$^*$\\
&G21&	50&	3&	3&	3$^*$\\
&D15&	62&	4&	4&	4$^*$\\
\hline
\end{tabular}
\end{center}
\caption{Match between original human coder consensus, ML model prediction and human re-coding consensus. *indicates consistency between ML prediction and re-code score.} 
\end{table}

\section{Conclusions and future work}
In this paper, we proposed a novel approach to get insights into student learning and outcomes based on a novel ML-NLP framework. Our results show that modern ML-NLP workflows may be effectively used as tools to support education research on student learning. The proposed method doesn't need any feature engineering and is able to work on a small dataset. The achieved QWK of $0.65$ is high and can provide a valuable feedback to the instructor. 

\section{Codes and Data Release}
Coming soon... 

\section{Acknowledgements}
We would like to thank Eric Miller, Mark Hempstead, and Kristen Wendell for helpful discussions. 

\bibliographystyle{unsrt}
\bibliography{ArXiv_Nov2020_Version1.bib}

\end{document}